# Estimation and Exploitation of Objects' Inertial Parameters in Robotic Grasping and Manipulation: A Survey

Nikos Mavrakis[a,∗], Rustam Stolkin[b]

[a]*Surrey Space Centre, University of Surrey, United Kingdom*
[b]*Extreme Robotics Lab, University of Birmingham, United Kingdom*

**Abstract**

Inertial parameters characterise an object's motion under applied forces, and can provide strong priors for planning and control of robotic actions to manipulate the object. However, these parameters are not available a-priori in situations where a robot encounters new objects. In this paper, we describe and categorise the ways that a robot can identify an object's inertial parameters. We also discuss grasping and manipulation methods in which knowledge of inertial parameters is exploited in various ways. We begin with a discussion of literature which investigates how humans estimate the inertial parameters of objects, to provide background and motivation for this area of robotics research. We frame our discussion of the robotics literature in terms of three categories of estimation methods, according to the amount of interaction with the object: *purely visual*, *exploratory*, and *fixed-object*. Each category is analysed and discussed. To demonstrate the usefulness of inertial estimation research, we describe a number of grasping and manipulation applications that make use of the inertial parameters of objects. The aim of the paper is to thoroughly review and categorise existing work in an important, but under-explored, area of robotics research, present its background and applications, and suggest future directions. Note that this paper does not examine methods of identification of *the robot's* inertial parameters, but rather the identification of inertial parameters of *other objects* which the robot is tasked with manipulating.

## 1. Introduction

Robots are being increasingly used in different environments that require object manipulation and task execution. In household environments, robots may need to grasp and manipulate everyday objects such as utensils, glasses, books, trays etc[1]. Robots

This work was funded by EU H2020 RoMaNS, 645582, and National Centre for Nuclear Robotics, EPSRC EP/R02572X/1. Stolkin was supported by a Royal Society Industry Fellowship.
∗Corresponding author

in production lines and warehouses manipulate objects such as components, mechanical parts and packaged goods [2][3]. Nuclear and disaster robots manipulate objects such as debris, bricks, and industrial waste materials [4][5]. Consequently, improving the efficiency and accuracy of grasping and manipulation algorithms remains an important and open area of robotics research. Detailed descriptions and approaches of robotic grasping and manipulation methods, as well as challenges in the field, can be found in [6][7][8].

In order to fulfil the assigned tasks, robots need to acquire useful information about the objects to be handled. This information is collected from the robot through various sensor and sensing modalities. Examples of such modalities are the visual and tactile perception streams of a robot system. Each modality requires specific sensing methods (equipment and/or algorithms) to gather raw data from the object. The raw data depend on the sensor used, and can be processed to extract different object properties. One example of such a sensing modality pipeline is visual perception, where a robot may use a depth sensor (sensing method), to extract an organised point cloud representation of the object (raw data) and process it to get the object's geometry and size (object information). Another example is auditory perception, where a robot can hit an object and measure the hitting sound frequency (raw data) gathered through a microphone array (sensing method), to extract the material density of the object (object information). In the physical world, a very important set of physical properties that characterise the object's motion are the *inertial properties*, namely the mass, centre of mass (CoM) and the inertia tensor. The inertial properties provide information on the object's heaviness and resistance to motion changes, as well as the mass distribution w.r.t. the object's volume. Knowing how an object will react during interaction and handling can make robotic grasping and manipulation more intelligent and efficient. However, one important issue with the inertial parameters is the lack of sensing equipment to directly identify them, as well as the absence of well-defined raw data that describe them (i.e. the analogue of image to the visual properties). Consequently, this lack of a sensing method and raw data description, has led to limited exploitation of the inertial parameters in robotics.

In literature from mechanical engineering, there are numerous existing methodologies on how to estimate these parameters, and they are presented in [9]. Usually, these methods require special hardware and execution (e.g. vibrating tables to measure the oscillation modes), and are ideal to use in more controlled environments, such as some industrial plants. As a result, they can be difficult to transfer in robot environments, as such hardware may not be available. While some basic principles are shared between the two fields, estimation in robotics is usually conducted by approximating the physical laws and relations that include the inertial properties (e.g. accelerating motion) and inference.

The main goal of this paper, is to present and categorise methods for estimating the inertial parameters of objects in robotics fields, and showcase their potential in robot grasping and manipulation. In this review we focus only on rigid objects. In addition, we do not focus on how the robot's estimate *their own* inertial parameters, i.e. the dynamics of their mechanical links, but we are interested on estimation methods of *other objects*. We attempt to organise the presented studies in categories based on the estimation methods and the environments that they are suitable for. We wish to demonstrate which sensing modalities are used for the identification of the inertial parameters and what raw data can be extracted for processing by the estimation algorithms. We describe how the estimation is conducted in model-based ways, and how state-of-the-art methods are incorporating novel data-driven algorithms. To the best of our knowledge, this is the first attempt to survey the field of inertial parameter estimation in robotics, and demonstrate its' usefulness by providing a large number of works that use them in grasping and manipulation.

The rest of the paper is organised as follows: In Section 2, a brief description of how humans perceive and use the inertial parameters is given, to provide a solid background and motivation. Section 3 provides the techniques in robotic estimation of the object's inertial parameters, and categorises them in three categories. Section 4 describes how the inertial parameters are used to solve some common problems in robot grasping and manipulation. In Section 5, challenges for the estimation and usage are mentioned, and in Section 6 a conclusion of the paper is given.

## 2. Human Perception and Exploitation of Inertial Parameters

Motivation for the use of inertial parameters in robotics comes from the corresponding human perception of object's inertial parameters and heaviness. While there is a large variety of psychophysics works that study this domain, we mention some prominent ones to provide some context behind the robotics works.

Humans tend to feel that larger objects are heavier than smaller objects of equal mass. This phenomenon is called the *size-weight illusion*, and has been studied for over a century [10][11][12][13]. The illusion is multi-modal [13], and it has been confirmed to appear even when a human has some prior knowledge of the object's size, be it visual or tactile cues [14]. While it it difficult to exactly pinpoint the origin of the illusion, the mass and inertia tensor of the object have been shown to influence it [12][15].

Mass in nature is realised in two ways: gravitational (static) mass and inertial (dynamic) mass. Gravitational mass is felt when an object lies on a person's hand, and inertial mass is felt when the person alters the motion state of the body. Humans perceive mass in one of the two ways, or a combination [15]. The two masses are equivalent in physical sense. Nevertheless, the authors in [16] showed that inertial mass perception of humans is highly affected by the motion type (acceleration or deceleration) and magnitude, and that the dynamic mass perception result can be two times lower than the static. These results are in accordance to previous experiments done in weightless environments [17]. In a recent approach [18], the inertial mass perception of humans that push a trolley was studied, and a linear mixed model was generated from a large number of reference stimuli. This linear mixed model can be transferred to power amplifying systems (such as robotic devices) to assist in human-robot collaboration tasks. Even without interaction, people are able to accurately infer relations between masses of objects in a scene through a mental simulation of the objects' interactions [19].

In [22], it was demonstrated that the estimation of a planar object's CoM from a pinching grasp, varies with the object's shape, size, symmetry and orientation.

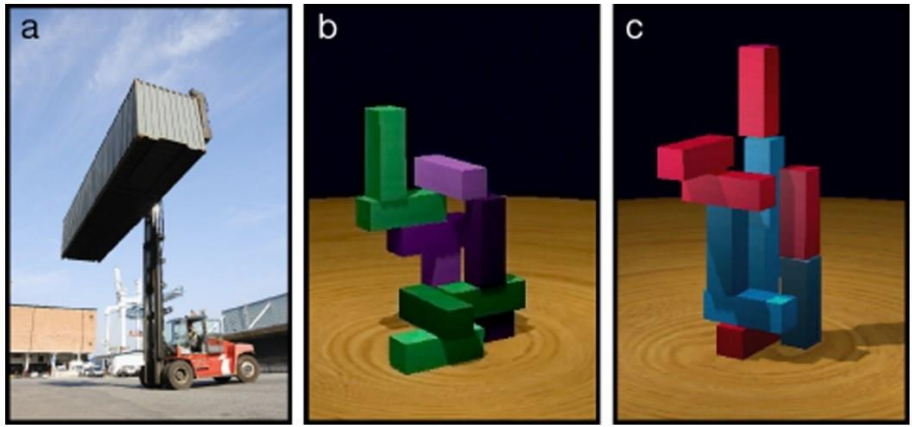

(a)

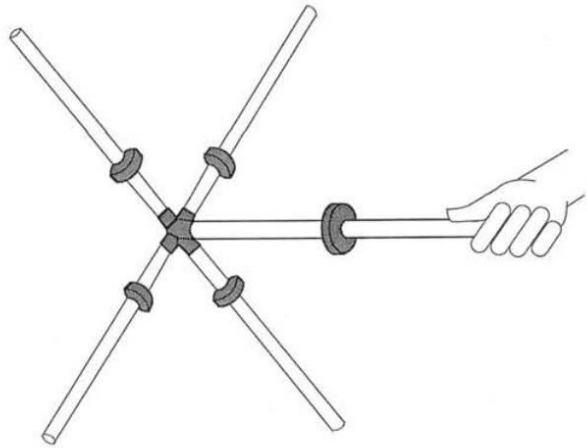

(b)

Figure 1: Numerous studies have shown how humans reason about the inertial properties of objects and how they involuntarily use them for perception. (a) Using complex scenes like the pictured, the authors in [19] demonstrated that humans are capable of inferring masses of objects by playing a mental simulation of the expected scene outcome, if given sufficient prior information. The inferences have been shown to be quite accurate. (b) A *tensor object*, is a set of cylinder handles with adjustable ring weights. As the rings' positions are adjusted, the object's inertia tensor changes. Studies have used such objects to demonstrate that a person's perception of an object's properties such as length, orientation in space, and grasping point on it, are a function of the object's inertia tensor [12][20][21].

Finally, there are numerous studies that show how humans feel properties of held objects by using their mass distribution (i.e. elements of the inertia tensor) through a stimulation mechanism of their muscular and tendon system, called *dynamic touch*. Through dynamic touch, people are able to estimate properties such as object orientation [20], position of a grasp relative to the object [21], and object length [23].

Since humans are good at grasping and manipulating objects, it is natural that the study of the perception and role of the inertial parameters in human manipulation, has led to applications in the robotics field. In the next sections, we provide an extensive list of works that showcase how robots can estimate and use the inertial properties of objects.

**3. Estimation of Inertial Parameters**

In this section, we categorise the estimation methods based on the amount and method of interaction with the object. We present three categories, *Purely Visual*, *Exploratory*, and *Fixed-Object*.

*3.1. Visual Methods*

The inertial parameters of an object are a function of their volume, volumetric distribution, and density. While the volume of the object is relatively easy to measure from visual cues, the density distribution throughout the volume is a property that is usually not known, or very difficult to determinate. Furthermore, it can be variable along the volume of the object. As a result, estimating the inertial parameters purely from visual elements seems like an ill-posed problem.

To solve this problem, the first attempts to estimate the inertial parameters had to assume known and uniform object density. Many objects satisfy this assumption, like most natural resources (e.g. wooden trunks, rocks), industrial components (e.g. wooden pallets, metal components, debris), and some household objects (e.g. dishes, chairs). Such an assumption simplifies the problem, and reduces it to extraction of volumetric properties. In [24], the authors calculate the inertia tensor of voxel representation of objects, assuming known and equal mass and size for each block, thus uniform density over the object. The inertia matrix and principal axes are calculated, and the octree/quadtree representation is projected on the axes to be used for recognition of 3D object models. One of the most famous approaches is in [25], where the author separated a rigid body into polyhedra of uniform density, calculated the mass, CoM, and inertia tensor for each polyhedron, and combined them to get the inertial parameters of the total object. The calculation of parameters was conducted by projecting the 3D volume integrals necessary for the identification of the inertial parameters as 2D surface integrals, and then as 1D line integrals using the divergence, projection and Green's theorem respectively. The author achieved fast computation times, and this method has been widely used in computer graphics.

In the first studies, density was used not as a property to be calculated but merely as a relationship that connects visual and inertial properties. As computing capabilities and image processing techniques improved, the relation between visual and inertial properties could be built from real data and statistical modelling. This is typical when the object is of organic nature, an animal, or an industrial component. In [26] the authors used a stereo camera pair to detect the geometric outline and truss measurements of swimming fish. These measurements have been proven to relate linearly with the fish mass, thus transforming the problem to shape detection. They were able to build a regression model that captures these linear relations, and tested the method with new fish populations with accurate results. Similar results were drawn in [27], where image processing techniques were used to calculate the volume of citrus fruits and a regression model was fit in their respective masses, and also in [28], this time using axis-symmetry of fruits for better image processing. Similar techniques have been employed for estimating the weight of livestock animals such as pigs [29] and chicken [30] for monitoring, and an analytical review on weight estimation of livestock animals can be found in [31]. In another approach, the authors in [32] estimated the weight of cup produce objects by extracting their volume in high-speed images, sampling and weighing some items in the produce, and using the measured weights to build a model to connect volume and weight.

Table 1: Overview of the inertial parameter estimation methods

| Method Type | Description | Pros/Cons | Estimated Parameters |
|---|---|---|---|
| Purely Visual | Use of only visual information (object geometry, RGB images, depth images, point clouds, video segments etc.) and possibly existing relationships between visual and inertial properties (density, size-mass formulas in organic objects etc.) | + Require little hardware<br>+ Easier to implement<br>+ Require easily obtainable raw data<br>- Require a lot of prior knowledge about the object<br>- Training can be time consuming and require large datasets | Mass, 3D inertial parameters (under assumptions) |
| Exploratory | Require basic interaction with the object. The applied forces and object motion are measured and the parameters are estimated from physical laws, or learning models. | + Accurate estimation<br>+ Ideal for most autonomous robotics scenarios<br>- Estimation based mostly on analytical models that require controlled environment | 2D inertial parameters, 3D inertial parameters by object tilting |
| Fixed-Object | The object is fixed on the robot's end-effector. As it moves along a trajectory, the endeffector wrenches and joint motions are measured and the parameters are estimated from the dynamic equations in the least squares way. | + Very accurate estimation<br>+ Portability through various robots<br>- Require grasping or fixing of the object, which may not be always possible | 3D inertial parameters |

The last decade has seen a sharp increase in computing capabilities, as well as sharp growth of big-data learning techniques. In the field of visual estimation of inertial parameters, there has been a shift in the research interest towards using big-data techniques for estimation. A notable case is presented in [33]. The authors used a large dataset of images taken from objects sold in Amazon.com, as well as the corresponding masses. They also generated a test set of household objects. They proposed a network architecture where the density of the object is calculated using an RGB image and a thickness mask, and the volume of the object is calculated from the object's bounding box and its' occupancy percentage. The two values were combined, and the mass estimate was provided. They compared the network performance with classical learning algorithms, as well as mass predictions from people in a human experiment. The result was minimum estimation error among other learning algorithms, and close performance to human intuition.

Another recent direction is using big-data for learning the underlying mechanisms of object interactions from visual data, and inferring the mass or relative heaviness of the objects in the scene. Such methods are typically trained by video sequences in simulation, and deployed on real sequences. Learning and modelling the interactions between objects is a very new and rapidly expanding field, and thus documenting all the related work is out of the scope of the paper. Instead, we provide a number of prominent works that are mostly related to learning interaction dynamics and include objects' inertial parameters in the procedure. The idea of describing object interactions as learning models originated in [34], where an *Object Oriented Markov Decision Process* was presented as a representation of object states and object interaction relations. The authors in [35] extend this notion, by introducing *Physics-Based Reinforcement Learning*, where the state dynamics and transitions are described to closely represent state-space Newton Euler dynamics. The uncertainties in object inertial and other parameters are modelled as belief over prior distributions. One of the most prominent studies is the one presented in [36]. In this paper, the authors presented Galileo, a learning model that is able to perceive physical properties of objects from video segments. It consists of a generative model that employs a physics engine to simulate object collisions, with an object-based hypothesis space. The generative model labels real video data by calculating the likelihood between the object velocity measurement in the real video and measurements from the physics-based videos. The labelled data are used to train a deep learning network, that can then estimate relative masses between the two objects, as well as other interaction outcomes. This work was extended in [37], where the authors employed physical laws to learn more object properties from unlabelled videos.

In [38] the authors introduced the concept of *Neural Physics Engine*. It also uses object-based representations of a scene, namely state vectors with each object's mass, friction and motion variables. A system comprised of an encoder, an interaction neighbourhood mask and a decoder are then able to predict the velocity of an object from previous states. The system was tested in inferring the object's mass based on a prior, and outperformed other well-known prediction methods.

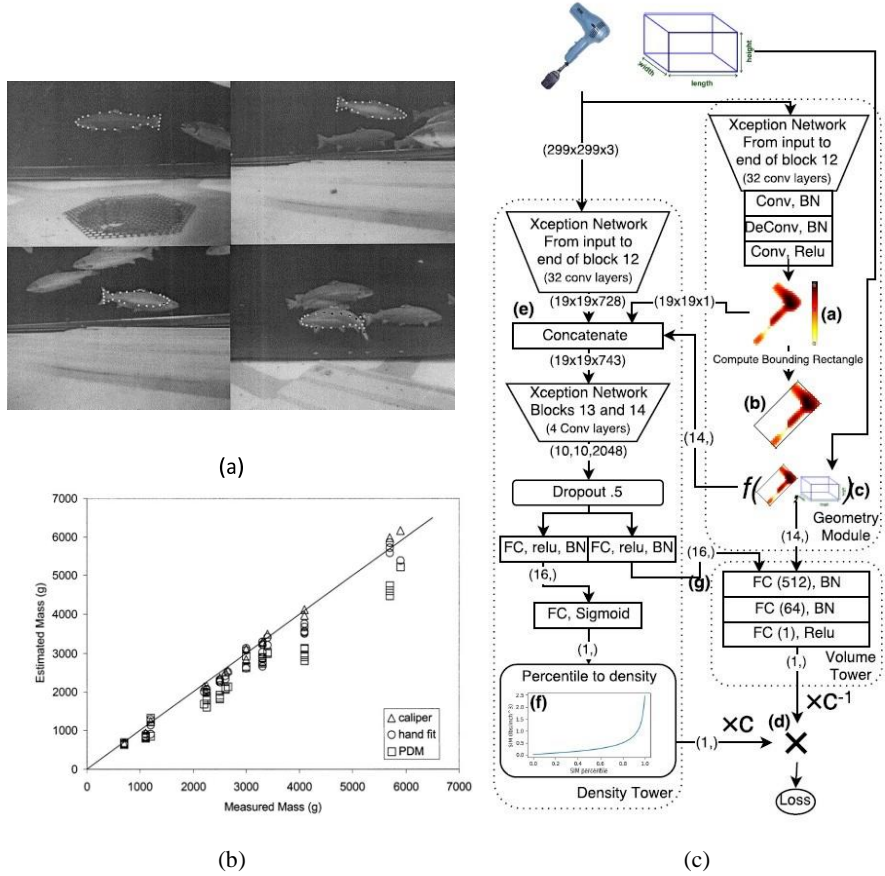

Figure 2: Mass estimation using purely visual features. When a physical connection between visual input (such as size, volume and shape) and mass exists between the object, the problem reduces to visual detection of features. (a) Extraction of visual features in video sequences, that correspond to the shape of the fish [26]. (b) When the extracted features are matched with measured masses, regression methods can be used to estimate the mass of new fish. A comparison of the estimated and the actual mass of the fish demonstrates the accuracy of such methods. (c) A state-of-the-art learning network for estimating an object's mass from a 2D image [33]. Two different network modules calculate the two elements needed for mass calculation: volume and density. They get as input an RGB image of the object, a thickness map and a bounding box. The network is able to calculate the object's mass almost as accurately as human perception. Novel learning approaches like this can be used to solve the ill-posed problem of estimating the inertial parameters from visual cues.

*3.2. Exploratory Methods*

As discussed above, a relation between visual and inertial properties (either density or other) is essential for purely visual identification of the inertial parameters. In autonomous robotics scenarios this information is usually not available, as an object of the environment may have non-uniform density, or be composed of objects with different density distributions. In this case, the robot needs to interact with the object in order to extract measurements useful for the estimation process. In this category, we include the works that require an amount of basic interaction with the object for the estimation, such as poking, pushing or tilting.

The inertial parameters of objects dictate their physical motion under the application of a force, through analytical physical laws. As a result, a classic approach for identifying the inertial parameters is by applying some simple way of contact on the object (pushing, poking, tilting, etc.), measuring the object's motion (velocity, applied force etc.) and calculating the inertial parameters from physical law equations. The first studies in the domain relied heavily on estimating the results through analytical laws of motion, and so required strong assumptions about the robot environment as well as complete control over the interaction procedure. For example, when a robot pushes an object on a surface and measures the motion, it needs to know the friction coefficient between the object and surface, the friction coefficient between the pushing finger and the object, as well as prevent the finger from sliding on the object's surface. A great deal of studies have been conducted for similar estimations, some with strict assumptions and some with more relaxed.

In [39], the authors determined the centre of friction of an object lying on a surface, by pushing with a mobile manipulator. They assumed the object's supporting surface to be a 2D square grid, and estimated the centre of friction and friction distribution over the grid using applied force and torque measurements. The pushing mechanics that were analysed in [40], [41] and [42] suggest that when the surface friction is uniform and isotropic, and the pressure distribution of the object's weight on the surface symmetric, then the centre of friction coincides with the 2D projection of the object's CoM on the surface. As a result, the work in [39] can be considered as an estimator of CoM under special circumstances. In [43] the author presented a series of methods of estimating material properties of objects with robot interaction. Among others, one of the presented methods suggested applying a quick strike on the object, measuring the applied force and its duration, and observing its velocity through a high speed camera. Then, through the impulse equation, the mass of the object can be determined. As the paper mostly described robot perception of material type through striking and listening to the resulting sound, no results were provided for the mass estimation method. Nevertheless, such an approach is plausible, especially nowadays with the existence of high rate force sensors and motion trackers. The authors in [44], were able to determine the mass and CoM of large (graspless) objects under limited knowledge of their shape. The object was high enough to be able to tilt. When the object was tilted, a *Gravity Equi-Effect plane* was defined as a plane between the CoM and the contact axis between object and surface. They proposed that since different planes intersect at the CoM, if three gravity planes are found, then the CoM can be estimated. By tilting the object, computing the planes from the object's geometry, and using the distance between finger and gravity plane, they were able to calculate the mass and 3D CoM of the object. They extended this method for the case of round-edged objects in [45], with similar results.

In [46] and [47] the authors estimated the mass of symmetric objects on a table. They applied a force on the visual centroid of the object, that started with low values and increased slowly. By detecting the force value that made the object move and comparing it with the normal force (that is a function of the static friction coefficient and the object's mass), they were able to calculate the object's mass. This work is one of the pioneering in inertial property estimation from pushing, and it required a lot of assumptions (symmetric object, known friction coefficient, non-sliding contact). Nevertheless, the estimation results were very accurate. In a similar configuration, the authors in [48] used a 2-fingered robot manipulator with force sensors to estimate a rectangular object's mass, 2D CoM and rotational inertia, as well as the motion friction coefficient. They applied a set of pushes on different sides of the object, and they measured the finger forces, as well as the motion of the fingers and object. The measurements were used to estimate the inertial parameters of the objects from planar Newton laws in a least squares approach.

The paper in [49] estimates the mass and rotational inertia of a pushed wheelchair. The authors use a simplified model of the pushchair's motion, and a special 2-fingered steering mechanism with 8 degrees of freedom. While pushing, the steering of the wheelchair was managed by an adaptive controller, that was able to provide stable measurements of the wheelchair's mass and rotational inertia after almost a minute of pushing.

Similarly to the Purely Visual category, the advances of image processing and statistical modelling, along with increased computing capabilities enabled the relaxation of the assumptions for complete control over the environment and interaction. The interaction could be studied as an action that carries noise and uncertainty. In [50], the authors estimated the CoM and mass distribution of an object, by applying a quick strike on pre-calculated candidate points. The object's point cloud and geometry, as well as the tumbling motion profile were used for the estimation of CoM, rotational inertia and mass distribution.

In [51] the authors proposed a decentralised approach, that uses a number of mobile robots pushing the object. Each robot pushes on a specific point on the object, and by using the applied forces and geometry of the pushing points that is communicated from each robot through a consensus algorithm, the authors are able to calculate the rotational velocity of the object, and thus the rotational inertia, 2D CoM, and mass. This approach is extended in [52], where the authors measure only the noisy velocity signals of each robot's end-effector. The authors in [53] conducted an identifiability analysis for the inertial parameters of objects with known geometry, when the objects are under sticking or sliding contact with a rigid surface. By expressing the motion equation of an object with contact and friction constraints as a complementarity problem, they demonstrated that with the object motion trackable, when the contact forces are unknown the identifiable inertial parameter is the mass-to-rotational-inertia ratio, and under known contact forces both mass and inertia can be identified. They confirmed this hypothesis in a series of experiment where they let objects with known geometry slide in a 2D plane and hit a surface. The formulation of this work was extended thoroughly in [54], with more objects. The authors in [55] estimated the inertial parameters of a large heavy object by applying a set of handling operations recursively (pushing, lifting, tilting), measuring the contact wrench and position as well as the object's pose, and calculating the likelihood in a grid-based Bayesian estimation scheme. An example from teleoperation is shown in [56], where the authors estimated the dynamic parameters of an object (mass and friction coefficients), to reconstruct a VR simulation useful for teleoperation. They did so by identifying three phases of the push depending on the object's motion, the static, critical and sliding phase. By modelling the contact forces as a mass-damper-spring, using different models for the frictional force in each phase, and measuring the object's motion, they were able to estimate the object's mass and build an accurate point-cloud-based VR environment that describes the scene.

Again, the cutting-edge techniques profit from the advancing of big-data and interaction learning techniques of the last decade. In these techniques, the physical laws are not used directly in the calculation, and instead they are translated as learning models that result from a large number of robot-object interactions. The Physics-Based Reinforcement Learning of [35] was further augmented in [57]. The authors used a mobile robot that pushes objects and measures the applied wrench and object motion. By maximising the log likelihood of the motion over given prior distributions and introducing a penalty term for object's ending state, the authors were able to estimate the mass and other physical parameters of the object. Other examples are the works in [58] and [59]. In these two works, the authors generated objects with different dynamic parameters (mass and friction coefficient) in the Bullet physics engine, as well as motion profiles when a force is applied on them. When they applied a force on a real object, they used Bayesian optimisation and Entropy Search to identify the simulated object whose motion closely matched the real object's motion. They then used the motion of the resulting simulated model to predict the motion of the real object, with high accuracy. While the authors primarily focused on the prediction part and did not provide results for the mass estimation part, their work is one of the first to employ big-data methods on the inertial parameter estimation problem. Finally, in [60] the authors use a Bayesian Regression Model to learn the inertial parameters of a hospital walker, by tracking the motion of a real robot pushing the walker in 39 trials. They used the learned model for manipulation planning, prediction and control of the walker motion, achieving low errors.

*3.3. Fixed-Object Methods*

The third estimation category includes the methods where there is a fixed connection between the robot system and an unknown load. This connection manifests in many forms, i.e. a robot that rigidly grasps the object, or has it otherwise attached at the end-effector. Studies in this category are inspired from classical dynamic model identification techniques in robotics, where a robotic arm executes excitation trajectories and gets torque measurements from the joints and force measurements from wrist sensors. The model can then be identified by rearranging the dynamic equations and solving them in a least squares way. Examples of such work include [61][62][63], and a survey paper in [64]. We mention only a couple of these studies as our main focus is the determination and usage of the inertial parameters of *other objects* and not the robot itself, but the techniques are similar in both cases.

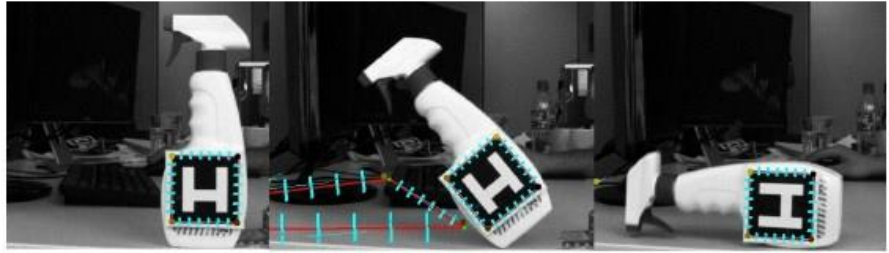

(a)

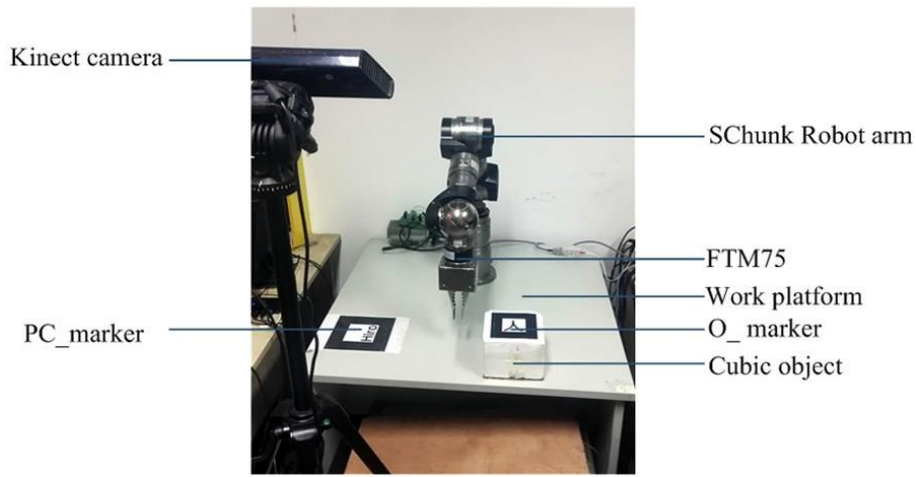

(b)

Figure 3: Examples of exploratory methods. By applying a simple action on the robot and observing its motion, the inertial parameters can be calculated. (a) The mass, CoM position and mass distribution can be calculated by striking an object, measuring the applied force, and tracking its rotational motion profile. Image from [50]. (b) Similarly, the object can be pushed by a robot and the inertial parameters can be estimated by planar motion laws. Image from [56].

Unlike the other categories, in this category all the 3D inertial parameters of the object are usually calculated, due to the capability of object motion in 3D space. Traditionally, these methods were preferred and optimised for robots operating in industrial environments. One of the first works in this category was the study in [65], where the authors proposed the equations for estimating the dynamics of a robot with rotary joints as well as the dynamics of the load, by using measurements of joint angles, velocities, accelerations, torques, and force and torque applied on the load. They proposed test motions of only one joint at a time, and proposed the equations for estimation, without conducting an experiment. In [66], the authors provided a method for estimating the parameters by using only measurements of force and torque on a wrist, as well as linear and angular position, velocity and acceleration of the sensing frame. They conducted two experiments with two different manipulators, and concluded that the estimation is more accurate when the measurement signals are less noisy. The authors in [67] identified the inertial and frictional parameters of both the robot and the load by employing integral dynamic model, which is a function of only the joint and load positions and velocities, and not accelerations. The identification was done by measuring force and torque on the wrist, as well as joint positions, velocities and torques, and applying a least squares estimator between measured and calculated dynamics. A similar approach was taken in [68], where the authors determined the 3D dynamics of a load using a maximum likelihood method, after getting noisy joint torque measurements and noise-free joint motion measurements. In addition, they were able to identify uncertainties in the robot dynamics, such as motor and transmission losses. They tested the model in an industrial manipulator with accurate results. In [69], the authors used force and torque measurements from a sensor on the wrist of an industrial robot, as well as measurements of angular velocity, and linear and angular accelerations, to estimate the inertial parameters of the load. They used excitation trajectories and applied a Total Least Squares method for the estimation, which enabled them to make on-line predictions on the parameters. The method was able to provide an accurate estimation in as little as 1.5 seconds. The authors in [70] developed a system of estimating an object's mass and rotational inertia, while held by a robot and being moved in a pendulumtype oscillation way. In [71], the authors developed a method similar to the ones in [66] and [69], with the difference that they used the torque difference between motion based on the calculated dynamics without the load and actual motion. To separate the effect of the mass, CoM and inertia tensor on the motion, the authors used 3 different excitation trajectories. In [72], the authors used a Weighted Least Squares algorithm, with positive-definite constraint on the inertia matrix. Finally, the authors in [73] used an Extended Kalman Filter to calculate the motion accelerations of an industrial robot with a load, and a Recursive Total Least Squares method combined with wrist force and torque measurements for the identification of the load parameters.

Recently, here has been a trend of taking robots out of the factory cages and operating them in close proximity of humans, as well as in outdoor environments. In addition, unmanned aerial vehicle (UAVs) manipulation is an emerging research field. These trends have shifted the focus of fixed-object inertial estimation from a single industrial arm, to UAVs, multiple robots, or even human-robot interactions. The core of the estimation process remains the same i.e. measurements of motion and force signals and model-based estimation of the parameters. In [74], the authors designed an unmanned aerial vehicle (UAV) with two customised grippers, and provided the Euler equations of motion for the UAV holding an object. They estimated the mass, inertia and CoM during both undisturbed hover and motion with disturbances. Similarly, the authors in [75] described the combined dynamic equations of a hexacopter with a 2-DOF arm. They demonstrated how the inertial parameters of an unknown load are incorporated in the total system dynamics, and how the inertial parameters of the object can be found from the known dynamics of the robot, and the motion of the hexacopter. They built an adaptive controller that can control the arm's end-effector on a desired trajectory, while estimating and compensating for the unknown load. A different approach was taken in [76], where the authors created an algorithm for the estimation of an object's mass and 2D CoM, by deploying a number of UAVs to lift the object. After iteratively deploying UAVs on the object and taking force measurements, an estimate was given by maximising the divergence between the force measurements and the parameters that produce the measurement. The authors in [77] used three mobile robots applying coordinated wrenches on an object, to estimate the mass, CoM, and inertia tensor of the object. They applied coordinated motions that resulted in pure translations or pure rotations on the object, and calculated the parameters with the grasping and motion equations of the composite system. The result was used to manipulate the object with minimum squeezing force. Estimation with coordinated transferring was also the subject of [78]. The authors estimated the inertial parameters of an object being handled by a robot and a human, by expressing the inertial parameters as a function of the robot and object motion. They then projected the robot motion used for identification in the null space of the grasp configuration around the object. This way, they were able to control the robot to match the desired human motion, and calculated the inertial parameters without disturbing the human task.

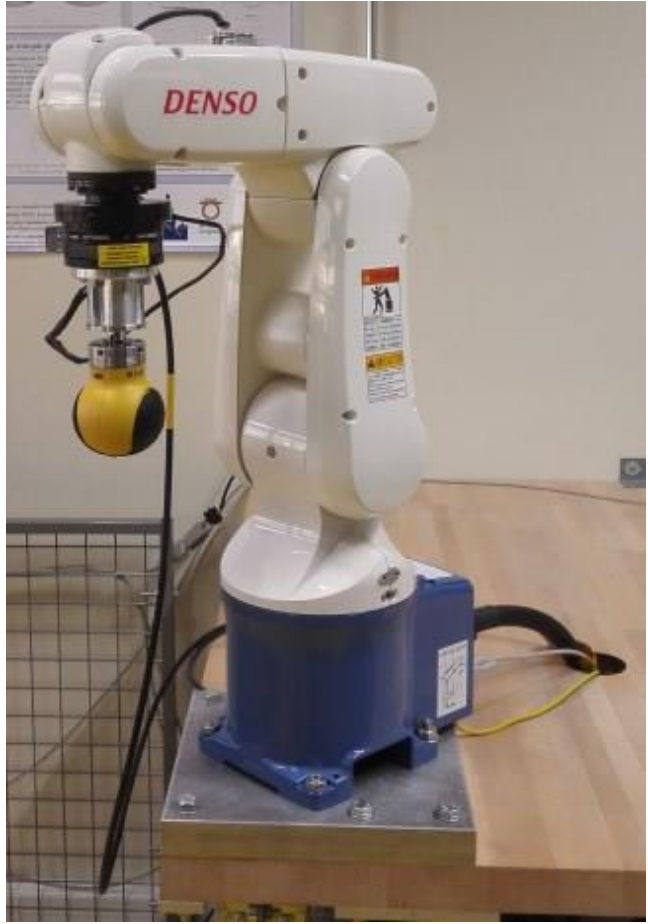

(a)

Figure 4: In load parameters identification, the load is grasped or otherwise attached to the robot and it is moved in the workspace. Motion, force, torque and and robot joint measurements are taken, and the parameters are estimated by using the equations of object motion and the robot dynamics, usually in a least squares way. For industrial applications, the load is fixed on the manipulator. The robot executes specific excitation motions that provide more informative measurements. By measuring the joint angles, positions, accelerations and torques, as well as wrist forces and torques, the load parameters are estimated from the total dynamic model. Image from [72].

Finally, machine learning has not been used extensively for this category, mostly due to the increased accuracy and efficiency of the model-based methods. In one of the first approaches to include robot learning in the procedure of load estimation, the authors in [79] simulated a 3-DOF manipulator to follow sinusoidal trajectories. They measured the joint positions, velocities, and accelerations, and used Locally Weighted Projection Regression to learn a dynamic model. By repeating this process with a number of simulated manipulators with reference loads attached, they were able to estimate the dynamics of new loads. In [80], the authors conducted a simulation to map changes in a held payload's mass, to variances in the joint torques. After simulating for different masses and motions, a set of clustering algorithms were compared to separate the payload variations into classes. These classes can then be used to distinguish between different masses that are fixed on the robot arm.

## 4. Exploitation of Inertial Parameters

In the last sections we described how state-of-the-art learning methods attempt to directly learn dynamic models from object motions generated by physics engines. This eliminates the need to estimate the object's inertial parameters, as the models that dictate the dynamic motion can be learned and represented in an intelligent way (e.g. a neural network or a statistical representation), and applied to many different objects without retraining. However, there are many cases where the value of the inertial parameters need to be determined numerically.

As discussed in previous sections, humans both intuitively and actively use the inertial parameters of objects to perceive their other properties, generate fixed and stable grasps on the object surface and facilitate manipulation tasks. For example, when a person encounters a heavy box, they realise that they must use both arms to lift it and transfer it. By nature, the person will place their hands on the object in an antipodal way, to provide support and stability, as well as put less strain on their arm muscles. When they transfer the object, they will keep their posture upright to minimise back strain. In this case, the person has conducted grasp and manipulation planning influenced by the object's inertial parameters. As humans are very effective in manipulating objects, it is natural that this ability to exploit the objects' inertial parameters should be transferred to robots as well.

In this section we present a large number of papers that demonstrate how the inertial parameters of objects are incorporated in algorithms to make robot grasping and manipulation more efficient. It is not easy to categorise these papers, as disproportionate amounts of work exist for different manipulation tasks. We mention works from the fields of grasp planning, manipulation planning, and controller design. The goal of this section is to show how the inertial parameters can be used in different grasping and manipulation tasks, and so justifying the need to estimate them in the first place.

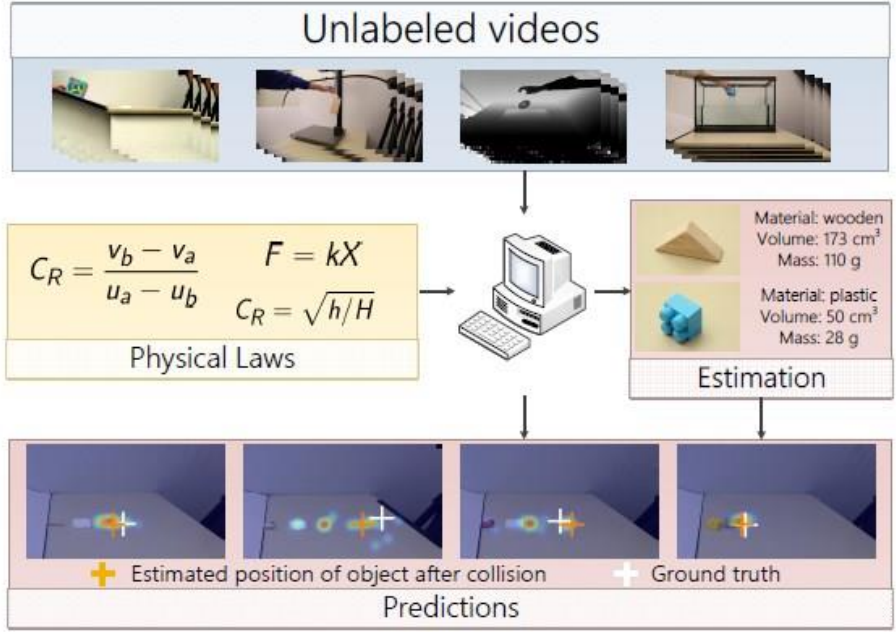

Figure 5: Recent estimation methods use large amounts of data to model dynamic interactions between objects. They then learn the mapping between video frames and motion, and use this mapping to infer the object's mass and friction coefficient. The system shown in the figure is able to infer physical properties of objects by being trained on video segments of their physical interactions. . Image from [37]

The object's inertial properties can be used to augment the dynamic model of a robot. Indeed, when a robot rigidly holds an object, the dynamic model changes according to the dynamics of the object and the grasping point on the object. One of the first works to provide analytical formulations for this property was in [81]. The author projected the dynamics of the object and the dynamics of the robot on a specific point in space, called *operational point*, and showed that the total dynamics can result from a simple addition of dynamic matrices. The formulation was also provided for the case of multiple manipulators handling an object. The results were extended in [82], where the dynamics of robots and objects were used to formulate other manipulation criteria such as reflected load on a manipulator, effective mass and inertia in manipulation movements and dynamic consistency. In addition, the augmented object model was used in [83] for formulating multiple-arm space robots that hold a captured object, as well as in [84] for branching configurations in which two robots with two arms each manipulated an object.

Our previous work has also used the augmented dynamics of robots and objects, in the novel field of task-oriented grasp selection. We evaluated a set of possible grasps on an object. Each grasping point results in different reflection of the object's inertial parameters on the operational point, and thus different augmented dynamics. We used the different combined dynamics to produce manipulation quality criteria that minimise the joint effort [85], minimising the felt impact force in case with a collision with the robot's end-effector [86], as well as to show how a grasp selection function could be used that satisfies these criteria [87]. Additionally, we used the combination of a space robot's dynamics and a captured object's inertial parameters to generate joint trajectories that minimise the applied strain on the object [88].

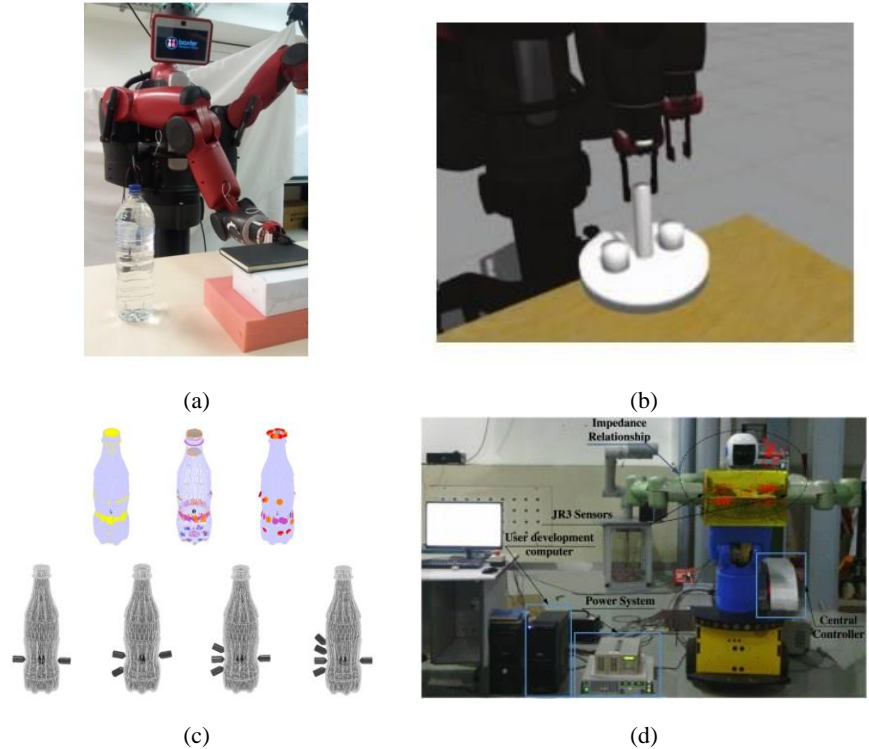

Figure 6: The inertial parameters of objects are used as a property that makes robot grasping and manipulation algorithms more efficient. (a) The robot is able to minimise the collision force by selecting different grasping points on an object. The selection criterion is based on the object's inertial parameters. Image from [86]. (b) Again, the robot uses the inertial parameters to select a grasp that will minimise the required joint torque to execute a task. Image from [85]. (c) Some grasp synthesis algorithms use the object's inertial parameters to generate stable and minimum-disturbance grasps. Image from [89] (d) The inertial parameters are also incorporated in the dynamic model of the robot for designing controllers, in this case a bio-inspired impedance controller for dual arm robots. Image from [90].

Most grasping algorithms in the literature exploit the geometry of objects to plan for and execute grasps. Nevertheless, the inertial parameters are essential in the analytic formulation of grasping, as both the grasp force mapping and the closed-form equations of motion require the dynamics of the object [91]. As a result, a lot of studies have used the inertial parameters of objects to solve problems in robot grasp synthesis, planning and quality evaluation. In [92], the author provided a grasp stability analysis that incorporates both spatial and contact stability. By including the inertial parameters of the object, the author developed a matrix, the eigenvalues of which are a metric for the grasp stability. In a similar manner, the authors in [93] developed a stability grasp metric by dividing the grasp stiffness matrix and the object's inertial matrix. Again, the eigenvalues of the division were related to the stability of the grasp. The authors in [94] developed a grasp quality metric by using the *Grasp Wrench Space*, namely the set of wrenches a grasp can counterbalance, and the *Object Wrench Space*, namely the set of wrenches an object will produce while moved for a task. For a given grasp, the scaling between the two spaces acts as a grasp quality measure. To construct the Object Wrench Space, the mass and CoM of the object are required. In [95], the authors generated antipodal 3-contact grasps on an object, by using the CoM, principal axis of inertia, and local object geometry. They argued that this type of antipodal grasp, where a grasping axis is aligned with one of the object's principal axis of inertia is intuitively more robust to gravity and accelerating forces. Similarly, the authors in [89] generated grasps on object models, by identifying *minimum inertia* regions on the object. These regions were defined as those where the contact point friction cones contained the object's CoM. By also exploiting the object's surface properties and the finger area the authors produced force-closed grasps. These results were also extended in [96], where the author selected the minimum inertia regions that were anthropomorphic, i.e. the finger contact normals were opposing or triangular-shaped. In [97], the authors created an efficient algorithm for the bin-picking problem for estimating the object's pose after it has been grasped, and the pose was determined by the on-line estimation of the objects principal axis of inertia and CoM. The authors in [98],[99] and [100] described the *sensation of grasping* of a robot, as a fuzzy inference method of the object's approximate size based on the object's principal axes of inertia. The sensation of grasping was used for an integrated robotic perception system that clears a table. In [101] the authors used a robot that grasps an elongated object from a graspable position close to the object's visual centroid by slightly lifting the object and measuring the torque signal on the robot's wrist, they were able to iteratively adapt the grasp towards the inertial CoM of the object.

Finally, other examples of inertial parameters exploitation include [102] where the authors used the inertial parameters in a bin-picking problem, to recognise objects and their poses for selecting good grasps, and [103], where the authors used the inertial parameters of heavy objects to define a strategy decision system for different manipulation types (push, lift etc.).

The inertial parameters of objects are included in in-hand dexterous manipulation and control studies. Example applications include object re-grasping, re-orientation and control of finger slippage. In [104] and [105], the authors provided the kinemodynamic modelling and controller design of a robot grasp on an object with rolling and sliding contacts respectively. For modelling and control they assumed knowledge of the object's inertial parameters. In [106] and [107], the authors further use gravitational torque and visual tracking with known object's inertia, to control the in-hand planar motion of an object while grasped by a pinch grasp. In [108] and [109], the authors examined the problem of using the environment contacts to manipulate an already grasped object, defined the dynamics of different prehensile pushing primitives by using the object's inertial parameters, and conducted experiments using simple objects. The authors in [110], described a finger motion planning scheme for in-hand manipulation of objects. They also used the inertial parameters of the object incorporated in the planning equations.

Finally, the inertial parameters of objects are necessary to develop algorithms for motion and control of dual or more manipulator robot systems. In parallel with the robot finger grasping case, the object's inertial parameters are incorporated in the dynamics of the robots to generate closed-loop models. The closed-loop models are used for solving manipulation problems such as control, trajectory generation, and others [111] [112][113][114] [90]. Referring to all literature that uses multiple arms and object models is out of the scope of the paper, and an extensive survey has been conducted in [115].

## 5. Discussion

The separation of estimation methods in different categories was conducted based on the nature and amount of robot-object interaction. Each type of robot-object interaction requires different data types for measurements. As a result, each estimation category is best suitable for specific environments.

Purely visual methods work best assuming some prior object knowledge. That makes them ideal in environments such as industrial plants, where components of known materials need to be inspected. By using only the object's geometry acquired by cameras or depth sensors, the system can extract the inertial parameters of the component and forward them to the next in line production steps. From there, the components can be handled (packed, transported etc.) more efficiently. Another possible use would be in robotic exploration, e.g. planetary, where the material composition of surrounding obstacles in a planetary body may be determined from spectral measurements, and the inertial parameters from geometry. The purely visual methods are, under strong assumptions, able to calculate all the inertial parameters of the object. They also require only visual data (images,depth maps etc) and limited equipment to operate. Nevertheless, they lose accuracy in case the object's density distribution is unknown.

Exploratory methods are suitable for autonomous robotics, where a robot does not have any information about the environment and surrounding objects. The estimation methods offer minimum levels of interaction and this makes them ideal to use in dangerous environments such as nuclear plants and disaster sites. A shortcoming of most methods in this category is the limitation of estimating only the 2D inertia parameters of an object. The equations of motion show that the object needs to be moved along all 3 coordinate axis to estimate its inertia tensor. By tilting along one axis or pushing on a planar surface, it it impossible to extract all the inertial parameters without prior knowledge. Nevertheless, mass and rotational inertia can be estimated and used as prior knowledge for uniform-density objects.

Fixed-object methods are often used in industrial plants to calculate the inertial parameters of heavier payloads. They require firm grasping or otherwise fixing of the object, which makes them ideal for controlled environments. They are quite accurate and able to estimate all the inertial parameters. Usually they require force sensors on the robot joints or the wrist.

As shown, the usage of the inertial parameters has been fundamental over robotics research in analytical robot grasping and manipulation. The algorithms presented need numerical values of the inertial parameters in order to work. This means that, despite the tremendous progress of the presented motion prediction and interaction learning models that learn dynamical models without explicitly estimating them, there is still a need for methods that can calculate the exact values of the inertial parameters.

## 6. Conclusion

In this paper we presented a number of classic and recent works that estimate the inertial parameters of objects, as well as characteristic usages of these parameters in robot grasping and manipulation. We separated the estimation works in three categories, and identified the advantages and shortcomings of each. Other categorisations can be made according to the data and estimation algorithms used. In addition, we believe that interesting research can come from combining methods. The progress in machine learning and big-data methods, can augment the estimation capabilities of exploratory methods to solve the ill-posed problem of extracting the 3D inertial parameters from a single push. Purely visual algorithms can provide an estimation prior for all other methods. The results from such combinations can be applied in numerous environments and further enhance the existing grasping and manipulation literature.